\pdfoutput=1
\documentclass[10pt, a4paper]{article}

\usepackage{lrec-coling2024} 

\usepackage{xcolor}

\usepackage{amsmath}
\usepackage{amssymb}
\usepackage{comment}

\usepackage{graphicx}
\usepackage{subfig} 
\usepackage{float}
\restylefloat{table}
\usepackage{graphics}
\usepackage{multirow}
\usepackage{multicol}
\usepackage{makecell}
\usepackage{arydshln}
\usepackage{booktabs}
\usepackage[font=small]{caption}
\usepackage[font=small]{subcaption}

\usepackage{enumitem}
\usepackage{pgfplots}

\title{EEE-QA: Exploring Effective and Efficient\\ Question-Answer Representations}

\name{
Zhanghao Hu*\quad
Yijun Yang*\quad
Junjie Xu*\quad
Yifu Qiu\quad
Pinzhen Chen
} 

\address{
School of Informatics, University of Edinburgh \\
\texttt{huzh666295@gmail.com\quad thomasyyj@outlook.com\quad smyjx1@163.com}\\
\texttt{yifu.qiu@ed.ac.uk\quad pinzhen.chen@ed.ac.uk}\\
}

\abstract{
Current approaches to question answering rely on pre-trained language models (PLMs) like RoBERTa. This work challenges the existing question-answer encoding convention and explores finer representations. We begin with testing various pooling methods compared to using the begin-of-sentence token as a question representation for better quality. Next, we explore opportunities to simultaneously embed all answer candidates with the question. This enables cross-reference between answer choices and improves inference throughput via reduced memory usage. Despite their simplicity and effectiveness, these methods have yet to be widely studied in current frameworks. We experiment with different PLMs, and with and without the integration of knowledge graphs. Results prove that the memory efficacy of the proposed techniques with little sacrifice in performance. Practically, our work enhances 38--100\% throughput with 26--65\% speedups on consumer-grade GPUs by allowing for considerably larger batch sizes. Our work sends a message to the community with promising directions in both representation quality and efficiency for the question-answering task in natural language processing.
\\ \newline \Keywords{Question Answering, Representation Learning, Memory Efficiency}
}

\begin{document}

\maketitleabstract

\def\thefootnote{}\footnotetext{*The first three authors contributed equally. Author ordering has been determined by coin flip.}
\def\thefootnote{\arabic{footnote}}

\section{Introduction}
Knowledge base question answering (KBQA) performs question answering (QA) using a knowledge base (KB) as its primary source of information \citep{mihaylov-etal-2018-suit, talmor-etal-2019-commonsenseqa, ijcai2021p611}. Popular KBQA approaches use a text encoder and a graph neural network (GNN) to derive representations for question-answer pairs and a knowledge base, followed by joint reasoning over their representations \citep{wang2022gnn,DBLP:conf/iclr/0001BYRLML22, sun-etal-2022-jointlk, hao-etal-2022-acenet, yasunaga-etal-2021-qa}.

Our motivation arises from two limitations in these works: (1) \textbf{inference efficiency}: the majority of QA models frame answer predictions as scoring to which degree a candidate answer can satisfactorily respond to a given question. Consequently, for each question, a model must perform inference as many times as there are candidates. We suggest an alternative approach that encodes all candidates alongside the question and trains the model to discern the most probable answer. This formalism enhances inference efficiency too since the QA model considers all candidates simultaneously in one go. (2) \textbf{semantic modelling for QA}: the PLM begin-of-sentence token, i.e. CLS, is commonly used as the representation for an input question-answer pair for {QA} prediction or {KG} interaction; however, it has long been argued that this representation is suboptimal in capturing the input semantics \citep{reimers-gurevych-2019-sentence}.

In this work, we first demonstrate that it is non-trivial for the QA model to infer a question with multiple candidates in a single pass. To address this challenge, we revisit the representation for input QA pairs and show that by simply pooling we can improve models' performance by a considerable margin. We then propose to delay the question-answer concatenation step to after a single-pass {PLM} encoding of the question and all answer candidates for improved memory efficiency. Experiments show that max pooling brings in substantial gains, even exceeding the current state-of-the-art KBQA models. On top of this, our proposed structure maintains a similar performance to the baseline while incurring less computation thus improving throughput. Our code is publicly available.\footnote{\url{https://github.com/Thomasyyj/EEEQA}}

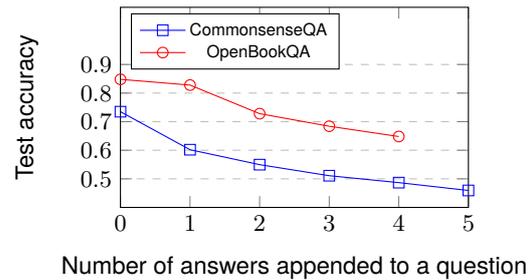
\begin{figure}[t]
\centering
\small

\begin{tikzpicture}
\begin{axis}[
    xlabel={Number of answers appended to a question},
    ylabel={Test accuracy },
    xmin=0, xmax=5,
    ymin=0.4, ymax=1.1,
    xtick={0,1,2,3,4,5},
    ytick={0.5,0.6,0.7,0.8,0.9},
    legend pos=north west,
    ymajorgrids=true,
    grid style=dashed,
    width=0.8\linewidth,
    height=0.55\linewidth,
    legend style={nodes={scale=0.7, transform shape}}
]

\addplot[
    color=blue,
    mark=square,
    ]
    coordinates {
    (0,0.7349)(1,0.6019)(2,0.5496)(3, 0.5109)(4,0.4867)(5,0.4593)
    };\addlegendentry{CommonsenseQA}
\addplot[
    color=red,
    mark=o,
    ]
    coordinates {
    (0,0.8480)(1,0.8280)(2,0.7280)(3,0.6840)(4,0.6480)
    };\addlegendentry{OpenBookQA}
\end{axis}
\end{tikzpicture}
\caption{A pilot experiment of appending random answers to the question before performing QA. Tested on CommonsenseQA (5 candidates) and OpenBookQA (4 candidates) with GreaseLM.}
\label{fig:MCQA answer}
\end{figure}

\section{From One to Multiple Answers}
\label{sec:hard-problem}

Intuitively, having all answer candidates together with the question in a single read may help models capture the nuances between them so as to make a better choice. Illustratively, it would be easier to cope with 2) than 1) below:
\begin{itemize}
    \item[1)] Seeing ``\textit{What do you do if you're late for an appointment? Cancel the appointment.}'' {and separately seeing} ``\textit{What do you do if you're late for an appointment? Apologize and explain.}''
    \item[ 2)] Answering ``\textit{What do you do if you're late for an appointment? Cancel the appointment, or Apologize and explain.}''
\end{itemize}

However, most {KBQA} works focus on modelling a score for each QA pair without inter-answer considerations. This also leads to larger memory usage when a system is deployed, as the question needs to be repeatedly encoded with each choice. A straightforward way to let the model access the information from all other answer candidates is to add all answer candidates to the question at the encoding stage. We perform a pilot test by incrementally and randomly appending answers from the candidate set to the end of a question and then use GreaseLM \citep{DBLP:conf/iclr/0001BYRLML22} to perform QA in the conventional style. We leave the technical description to Section~
\ref{sec:question answer pair concatenation}. As Figure~\ref{fig:MCQA answer} shows, with more answer choices added to the question, the system performance decreases monotonically. This reflects that naively dealing with multiple answers at the same time for {KBQA} will undermine the judgment of a system. We are therefore motivated to search for more feasible solutions.

\section{Methodology}

\subsection{Preliminaries}
\label{sec:prelim}

With a question in discrete tokens $Q=\lbrack x_{1},x_{2},...,x_{|Q|}\rbrack$ and a set of $n$ answer choices $\mathcal{A}=\{A_1,A_2,...,A_n\}$, current methods use a {PLM} $f()$ to encode the question and each  $A_i$ as a pair $(Q,A_i)_{A_i \in \mathcal{A}}$ into representations 
$\mathcal{H}=
\lbrack
H_{{\text{{\textless s\textgreater}}}},
H_{x_1},
...,
H_{x_{|Q|}},
H_{{\text{\textless}/\text{s\textgreater}}_1},
H_{{\text{\textless}/\text{s\textgreater}}_2},
H_{A_i},
H_{{\text{\textless}/\text{s\textgreater}}_3}
\rbrack
= f(
\lbrack
{\text{{\textless s\textgreater}}},
Q,
{\text{{{\textless}/\text{s\textgreater}}}}_1,
{\text{{{\textless}/\text{s\textgreater}}}}_2,
A_i,
{\text{{{\textless}/\text{s\textgreater}}}}_3
\rbrack
)$, where ${\text{{\textless s\textgreater}}}$ and ${\text{{\textless}/\text{s\textgreater}}}$ denote begin-of-sentence (or CLS) and end-of-sentence (or SEP) tokens respectively. This is carried out separately for each $A_i$ for a total of $n$ times. A multi-layer perception $g()$ then computes a score using the concatenation of the question CLS token and the answer embedding $S_i = g(H_{\text{\textless{}s\textgreater,i}} \oplus H_{A_i})$ for each $(Q,A_i)$ pair. The final answer $A_i$ is selected by $i = argmax(S_1,S_2,...,S_n)$. In this way, the PLM encodes the question with each answer candidate in $n$ passes, which we refer to as \textbf{$\mathbf{1}$A$\mathbf{n}$P} encoding.

\subsection{Improved question representations}
\label{sec: language pooling approaches}
In previous {QA} works, the question representation is used for the interaction with a {KG} (if included) and for score prediction, but the resulting $H_{{\text{{\textless s\textgreater}}}}$ from the conventional {CLS} pooling cannot capture the full semantics of the input \citep{reimers-gurevych-2019-sentence}. To seek a more meaningful representation, we propose a pooling operation $pool()$ over all question token embeddings. Having the same QA representations $\mathcal{H}$, the score for $(Q,A_i)$ can then be computed as 
$S_i = g(pool(
H_{{\text{{\textless s\textgreater}}}},
H_{x_1},
...,
H_{x_{|Q|}},
H_{{\text{\textless}/\text{s\textgreater}}_1}
)\oplus H_{A_i})$ 
instead. 
Our work investigates four different pooling operations:
\begin{itemize}[nolistsep]
    \item \textbf{Max pooling} selects the maximum element in each dimension from all embeddings.
    \item \textbf{Mean pooling} averages all embeddings.
    \item \textbf{Attentive pooling} sums up all embeddings, weighted by a learnable attention vector.
    \item \textbf{Layerwise CLS pooling} adds up all the begin-of-sentence embeddings from each {PLM} layer, weighted by a learnable attention vector \citep{tenney-etal-2019-bert}.
\end{itemize}

\subsection{Single-pass encoding}
\label{sec:question answer pair concatenation}

The conventional question-answer encoding is memory inefficient because the often long question needs to be encoded multiple times with different answers. We propose a one-pass technique, termed \textbf{$\mathbf{n}$A$\mathbf{1}$P}, where a {PLM} encodes a question and all candidate answers simultaneously to reduce memory usage. This also resembles how a human tackles a multiple-choice question: answers are compared to each other before a decision is made. We append all candidates to the question and embed with $f()$ in one pass:
$
\lbrack {\text{{\textless s\textgreater}}}, Q, {\text{{\textless}/\text{s\textgreater}}}_1, {\text{{\textless}/\text{s\textgreater}}}_2, A_1, {\text{{\textless}/\text{s\textgreater}}}_3,A_2...,A_n,{\text{{\textless}/\text{s\textgreater}}_{n+2}}\rbrack
$.

To alleviate the potential interference between answers when they are being encoded altogether as illustrated in Section~\ref{sec:hard-problem}, we propose two levels of semantic integration by comparing and fusing the answer representations and then merging them with the question. 
For the answer comparison, inspired by prior research which runs a gate mechanism over answers in multiple encoding passes \citep{Zhang_Zhao_Wu_Zhang_Zhou_Zhou_2020, hao-etal-2022-acenet}, we propose a gated interaction adapted to our single-pass scheme.

Specifically, we first perform the aforementioned pooling on the entire span of each answer token as its representation $H_{A_i}=pool(H_{{\text{\textless}/\text{s\textgreater}}_{i+1}},H_{A_i},H_{{\text{\textless}/\text{s\textgreater}}_{i+2}})$. For $H_{A_i}$ and all other answers $H_{A_j},_{j\neq i}$, we then compute a multi-head attention (MHA) score $\alpha_{ji}$ of $H_{A_j}$ queried by $H_{A_i}$ after pooling. Intuitively, this allows the explicit comparison between answer $A_i$ and each other answer $A_j$. 
We derive the final answer representation $\hat{H}_{A_i}$ by applying a gate mechanism to balance the information from answers $H_{A_i}$ and $H_{A_j}$ using the attention score $\alpha_{ij}$ computed as:
\begin{align*}
    \alpha_{ji} &= \text{MHA}(\text{query=}H_{A_i}, \text{key=}H_{A_j}, \text{value=}H_{A_j})\\
    \hat{H}_{A_i} &= \gamma H_{A_i} + (1-\gamma) (\textstyle\sum_j \alpha_{ji} H_{A_j})
\end{align*}
where $\gamma=\sigma(W_1 H_{A_i}+W_2 (\sum_j \alpha_{ji} H_{A_j}) +bias)$ is a weight to balance the information from a particular answer and other answers, $\sigma()$ is a sigmoid function, and $W_1, W_2$ are trainable matrices. Question-answer scoring remains similar to Section~\ref{sec: language pooling approaches}; we concatenate the pooling outcome of the question and each answer's gated representation to score each $(Q,A_i)$ pair: $S_i = g(pool(H_{{\text{{\textless s\textgreater}}}},
H_{x_1},...,H_{x_{|Q|}},H_{{\text{\textless}/\text{s\textgreater}}_1})\oplus \hat{H}_{A_i})$.

In addition, we formally define our setting in Section~\ref{sec:hard-problem} as \textbf{$\mathbf{n}$A$\mathbf{n}$P}, which affixes all multiple candidates to the original question to form an extended question, and then encodes the modified question with each candidate $A_i$ as a pair $(\lbrack Q,A_1,...,A_n\rbrack,A_i)_{A_i \in \mathcal{A}}$ into representations 
$\mathcal{H}$ for $n$ times separately for each $A_i$. The scoring criterion to select the final answer is similar to Section~\ref{sec: language pooling approaches}. This $n$A$n$P technique bridges the conventional approach and our single-pass scheme, and creates a fair comparison between $1$A$n$P and $n$A$1$P with controlled variations.

\section{Experiments and Discussions}

\subsection{Data and evaluation}
We evaluate our approaches on two QA datasets. First, \textbf{CommonsenseQA} \citep{talmor-etal-2019-commonsenseqa} contains 12,102 questions, each paired with five answer candidates; the questions demand commonsense knowledge. We follow the in-house data split by \citet{lin-etal-2019-kagnet} because the official test is not public. Next, \textbf{OpenBookQA} \citep{mihaylov-etal-2018-suit} comprises 5,957 4-way multiple-choice questions related to elementary scientific knowledge. For this dataset, we follow the official data splits. \textbf{Accuracy} (\%) is used as the metric.

\subsection{Experimental setup}

\paragraph{Pooling} We first explore the best pooling strategy based on GreaseLM \citep{DBLP:conf/iclr/0001BYRLML22} since it uses the CLS token for both KG interaction and QA scoring, but we consider this suboptimal. To position our pooling results in current research, we include several prior works for comparison. 

\paragraph{Efficient inference} We compare our proposed methods: single-pass inference $n$A$1$P, and the vanilla scheme $n$A$n$P from Section~\ref{sec:question answer pair concatenation} with the conventional $1$A$n$P. We extensively test them in three scenarios: PLM-only \cite[{RoBERTa-Large},][]{liu2019roberta}, PLM with KG \citep[PLM+KG,][]{yasunaga-etal-2021-qa}, as well as PLM with KG and an interaction node \citep[PLM+KG+Int,][]{DBLP:conf/iclr/0001BYRLML22}.

\paragraph{Hyperparameters} Our PLM backbone is RoBERTa \citep{liu2019roberta}. In experiments involving a KG, we use {ConceptNet} \citep{xu-etal-2021-fusing}. GNN node embeddings are initialized with entity embeddings from \citet{feng-etal-2020-scalable}. For each QA pair, we retrieve the top 200 nodes and their adjacent edges based on node relevance scores following \citet{xu-etal-2021-fusing}. We use a dimension of 200 and 5 {GNN} layers, with a 0.2 dropout probability. We use batch sizes 64 and 128 for CommonsenseQA and OpenBookQA correspondingly. We train all models with the {RAdam} optimizer \citep{liu2019variance} on a single Nvidia RTX A5000. Learning rates for the PLM parameters and non-PLM modules are 1e-5 and 1e-3 separately. We determined these hyperparameters using the development set. We note that we use a slightly different but necessary implementation in the PLM+KG+Int experiments with our $n$A$1$P framework since the information interaction requires an additional special token for each answer.

\subsection{Pooling results}
\label{results}

We first report the performance of our proposed pooling methods in Table \ref{Tab:pooling performance}. We see improvements in most pooling methods on the GreaseLM architecture, showing the usefulness of a finer representation. Among all, max pooling demonstrates an incredible gain, even outperforming current SOTA models. This highlights that a simple but effective representation has long been neglected in the community. We therefore stick to max pooling as opposed to CLS pooling for subsequent experiments.

\begin{table}[t]
\centering\small
\setlength{\tabcolsep}{-0.5ex}
\begin{small}
\begin{tabular}{lc}
\toprule
\multicolumn{1}{c}{\textbf{System}}  & \multicolumn{1}{c}{\textbf{Accuracy (std.)}} \\
\midrule
RoBERTa-Large \citep{liu2019roberta} & 68.69 ($\pm$0.56) \\
QA-GNN \citep{yasunaga-etal-2021-qa} & 73.41 ($\pm$0.92) \\
JointLK \citep{sun-etal-2022-jointlk} & 74.43 ($\pm$0.83) \\
ACENet \citep{hao-etal-2022-acenet} & 74.72 ($\pm$0.70) \\
GreaseLM \citep{DBLP:conf/iclr/0001BYRLML22} & 74.20 ($\pm$0.40) \\
GreaseLM (\citet{Ye_Cao_Chen_Xu_Zou_2023}'s re-run) & \hspace*{2.4ex}73.60 (unknown) \\
\hdashline
GreaseLM (our re-run)  & 73.57 ($\pm$0.08) \\
\hspace{1ex}+ mean pooling  & 73.73 ($\pm$0.29)\\
\hspace{1ex}+ max pooling   & \phantom{\textsuperscript{$\dagger$}}{75.42 ($\pm$0.52)}\textsuperscript{$\dagger$}\\
\hspace{1ex}+ attentive pooling & 73.97 ($\pm$0.51)\\
\hspace{1ex}+ layerwise CLS pooling & 73.97 ($\pm$0.16)\\
\bottomrule
\end{tabular}
\caption{Performance (\%, accuracy from 3 runs) of pooling techniques compared with previous works on CommonsenseQA. \textsuperscript{$\dagger$}Significantly better than GreaseLM and other pooling methods with p-values $<$ 0.05 in pairwise t-tests.}
\label{Tab:pooling performance}
\end{small}
\end{table}

\begin{table*}[ht]
\centering\small
\begin{tabular}{clcccccc}
\toprule
\multirow{2}{*}{\textbf{System}} & \multicolumn{1}{c}{\multirow{2}{*}{\textbf{Pooling}\phantom{poo}}} & \multicolumn{3}{c}{\textbf{CommonsenseQA}} & \multicolumn{3}{c}{\textbf{OpenBookQA}}\\
\cmidrule(lr){3-5}\cmidrule(lr){6-8}
 & & \phantom{A}PLM\phantom{A} &  \phantom{A}+ KG\phantom{A} & + KG + Int &  \phantom{A}PLM\phantom{A} &  \phantom{A}+ KG\phantom{A} & + KG + Int \\
\midrule
\multirow{2}{*}{\makecell{$1$A$n$P\\(previous)}} 
& CLS & 70.02 & 72.82 & 73.97 & 80.20 & 81.80 & 81.60 \\
& Max & 70.51 & 73.41 & 75.42 & 82.40 & 82.40 & 82.60\\
\midrule
\multirow{2}{*}{\makecell{$n$A$n$P\\(contrastive)}} 
& CLS & 67.12 & 69.62 & 68.82 & 79.40 & 78.40 & 81.80 \\
& Max & 67.12 & 68.90 & 69.14 & 79.40 & 82.60 & 82.20\\ 
\midrule
\multirow{3}{*}{\makecell{$n$A$1$P\\(ours)}} 
& CLS & 67.77 & 69.30 & 69.38 &78.80 & 79.40 & 80.20 \\
& Max & 68.25 & 68.65 &70.91 & 79.00 & 80.60 &80.40  \\  
& Max + Gate & 69.62 & 71.88 & 70.91 & 79.60 & 80.60 & 81.40 \\
\bottomrule
\end{tabular}
\caption{Performance (accuracy, \%) of our systems on CommonsenseQA and OpenBookQA.}
\label{Tab:main performance results}
\end{table*}

\begin{table*}[t]
\centering\small
\begin{tabular}{lcccccc}
\toprule
\multicolumn{1}{c}{\multirow{2}{*}{\textbf{\makecell{GPU\\Model}}}} & \multicolumn{1}{c}{\multirow{2}{*}{\textbf{\makecell{Mem.\\(GB)}}}} & \multicolumn{2}{c}{\textbf{Batch size ($\uparrow$)
}} & \multicolumn{2}{c}{\textbf{Inference time ($\downarrow$)
}} \\
\cmidrule(lr){3-4}\cmidrule(lr){5-6}
& & $1$A$n$P & $n$A$1$P ($\Delta$\%) & $1$A$n$P & $n$A$1$P ($\Delta$\%) \\
\midrule
RTX A5000      & 24 & 100 & 160 \phantom{1}(+60\%) & 4.61s & 3.31s (-28\%) \\
RTX 3090       & 24 & 100 & 160 \phantom{1}(+60\%) & 2.45s & 1.82s (-26\%)\\
RTX 2080 Ti    & 11 & \phantom{0}30 & \phantom{0}45 \phantom{1}(+50\%)  & 1.13s & 0.76s (-33\%) \\
GTX 1080 Ti    & 11 & \phantom{0}30 & \phantom{0}45 \phantom{1}(+50\%)  & 2.64s & 1.27s (-52\%) \\
Titan X Pascal & 12 & \phantom{0}40 & \phantom{0}55 \phantom{1}(+38\%)  & 3.56s & 1.55s (-56\%) \\
GTX 1080 & \phantom{0}8 & \phantom{0}10 & \phantom{0}20 (+100\%) & 2.21s & 0.77s (-65\%) \\
\bottomrule
\end{tabular}
\caption{Comparison in efficiency between $1$A$n$P and our $n$A$1$P: usable batch size and total inference time when solving 1000 QA instances on a single GPU.}
\label{Tab:time reduction}
\end{table*}

\subsection{From multiple to one pass}
The results from our memory-efficient experiments across two datasets are presented in Table~\ref{Tab:main performance results}.  
 
\paragraph{Pooling} For both CommonSenseQA and OpenBookQA, we observe a better performance with max pooling compared to CLS pooling in most of the settings, indicating the effectiveness and generalizability of pooling in QA representation. 
We note that max pooling is slightly behind when multiple answers are encoded simultaneously and when a KG is used. We conjecture that this is because, with many answers, it is difficult for the information in LMs to align with the corresponding KG sub-graphs.

\paragraph{Efficient inference} For our proposed $n$A$1$P encoding, results indicate that when seeking improved memory efficiency, the baseline CLS pooling accuracies on both CommonsenseQA and OpenBookQA are sacrificed to a slight degree. 
However, this can be mitigated through our proposed techniques: max pooling as well as the gated answer representation mechanism. We observe on par if not higher performance when these are added, highlighting the effectiveness of our explicit inter-answer interaction. 

\paragraph{Transferability} Additionally, we switch the {PLM} from RoBERTa to {BERT} \citep{devlin-etal-2019-bert} in order to investigate whether our $n$A$1$P framework is compatible with other PLM resources. We find that the test accuracies are 50.68\% for our $n$A$1$P approach and 49.80\% for the traditional $1$A$n$P. The pattern is the same as we have observed on CommonsenseQA. Nonetheless, the results with BERT as the PLM are significantly lower than RoBERTa.

\subsection{Throughput and inference time}
To study the practicality of memory efficiency, we run {GreaseLM} with both $1$A$n$P and our $n$A$1$P using the same configurations to solve 1000 QA instances. Table~\ref{Tab:time reduction} reports the maximum usable batch size and total inference time on a single consumer-grade GPU. To eliminate variable factors like input lengths, we duplicate a single data instance to fill up a batch.

Compared with the baseline $1$A$n$P, while the incorporation of the gated mechanism to our $n$A$1$P approach could introduce extra processing time and memory, we highlight both an increase in maximum batch size and a decline in inference time with our method. This framework enables the utilization of larger batch sizes across multiple {GPU} models which markedly increases throughput. This is particularly advantageous when conducting QA tasks at a large scale.

\begin{table}[ht]
\centering\small
\begin{tabular}{lc}
\toprule
\multicolumn{1}{c}{\textbf{System}}  & \textbf{Accuracy} \\
\midrule
Max Pool + Q$\oplus$A + Gate & {71.88}\\
Max Pool + Q$\oplus$A & 69.54\\
Max Pool  & {68.65}  \\
\bottomrule
\end{tabular}
\caption{Performance of our $n$A$1$P with ablations on CommonsenseQA.}
\label{Tab:ablation performance results}
\end{table}

\subsection{Ablation study}
For our proposed $n$A$1$P approach, we systematically analyze each its component's impact on the performance on the CommonsenseQA {in-house test}. We eliminate a single module at a time as shown in Table \ref{Tab:ablation performance results}. Introducing the QA concatenation yields a 0.89\% improvement, demonstrating the value of considering both questions and answer contexts for enhanced interaction and reasoning. Notably, having the gate layer module leads to a significant performance boost of 2.34\%, emphasizing the crucial role of diverse choice interactions during the message-passing process.

\section{Related Work}

Knowledge-base Question Answering necessitates models to jointly reason over both external knowledge graphs and parametric knowledge from language models \citep{talmor-etal-2019-commonsenseqa, mihaylov-etal-2018-suit}. However, grounding knowledge from both textual and structured modalities presents a non-trivial challenge \citep{yasunaga-etal-2021-qa, sun-etal-2022-jointlk, DBLP:conf/iclr/0001BYRLML22}. Furthermore, existing works predominantly adhere to the $1$A$n$P setting, which needs multiple inferences to answer a question \citep{liu2019roberta, hao-etal-2022-acenet, yasunaga2022deep}.

The only exception is a recent work using decoder-only large language models, where a question and all possible answers can be framed as a natural language input \citep{robinson2022leveraging}. We highlight that our multi-pass encoder models, with similar performance, are significantly more affordable in practice. Finally, Section~\ref{sec:hard-problem} shows that vanilla $n$A$1$P notably lags behind $1$A$n$P, which motivates us to revisit the representation scheme in current models.

\section{Conclusion and Outlook}

This paper presents our efforts in searching for an effective and memory-efficient representation for question-answering. 
Experimental results show that a simple max pooling mechanism consistently outperforms CLS pooling which has been widely used in this field. We then propose and test a gated single-pass inference approach to encourage answer interactions and improve efficiency. It is seen that the approach reduces memory usage with a tiny sacrifice in performance. Our paper calls for future work to shift to max pooling and encourages further exploration of efficiency-aware QA representations.

Our investigation focused on the PLM side of question answering, but the KG side remains unoptimized. In our experiments, sub-KGs are constructed for each QA pair, without adjusting for a single-pass encoding scheme where all answers are encoded simultaneously. Thus, although we could encode a question with all answers in our proposed $n$A$1$P inference approach, processing separate sub-KGs would slow down the inference. The challenge exists in combining all answer candidates' information into one knowledge graph which can be looked at by future research.

\section*{Acknowledgements}
This work originated from a course project at the University of Edinburgh. We must thank the course organizers and teaching staff. We are grateful to the anonymous reviewer who identified an oversight in Section~\ref{sec:prelim}.

Pinzhen Chen has received funding from UK Research and Innovation under the UK government's Horizon Europe funding guarantee [grant numbers 10039436 and 10052546]. 

\section*{Bibliographical References}
\bibliographystyle{lrec-coling2024-natbib}
\bibliography{lrec-coling2024-example}

\begin{thebibliography}{20}
\expandafter\ifx\csname natexlab\endcsname\relax\def\natexlab#1{#1}\fi

\bibitem[{Devlin et~al.(2019)Devlin, Chang, Lee, and Toutanova}]{devlin-etal-2019-bert}
Jacob Devlin, Ming-Wei Chang, Kenton Lee, and Kristina Toutanova. 2019.
\newblock \href {https://doi.org/10.18653/v1/N19-1423} {{BERT}: Pre-training of deep bidirectional transformers for language understanding}.
\newblock In \emph{NAACL}.

\bibitem[{Feng et~al.(2020)Feng, Chen, Lin, Wang, Yan, and Ren}]{feng-etal-2020-scalable}
Yanlin Feng, Xinyue Chen, Bill~Yuchen Lin, Peifeng Wang, Jun Yan, and Xiang Ren. 2020.
\newblock \href {https://doi.org/10.18653/v1/2020.emnlp-main.99} {Scalable multi-hop relational reasoning for knowledge-aware question answering}.
\newblock In \emph{EMNLP}.

\bibitem[{Hao et~al.(2022)Hao, Xie, and Zhang}]{hao-etal-2022-acenet}
Chuzhan Hao, Minghui Xie, and Peng Zhang. 2022.
\newblock \href {https://doi.org/10.18653/v1/2022.emnlp-main.579} {{ACEN}et: Attention guided commonsense reasoning on hybrid knowledge graph}.
\newblock In \emph{EMNLP}.

\bibitem[{Lan et~al.(2021)Lan, He, Jiang, Jiang, Zhao, and Wen}]{ijcai2021p611}
Yunshi Lan, Gaole He, Jinhao Jiang, Jing Jiang, Wayne~Xin Zhao, and Ji-Rong Wen. 2021.
\newblock \href {https://doi.org/10.24963/ijcai.2021/611} {A survey on complex knowledge base question answering: Methods, challenges and solutions}.
\newblock In \emph{IJCAI}.

\bibitem[{Lin et~al.(2019)Lin, Chen, Chen, and Ren}]{lin-etal-2019-kagnet}
Bill~Yuchen Lin, Xinyue Chen, Jamin Chen, and Xiang Ren. 2019.
\newblock \href {https://doi.org/10.18653/v1/D19-1282} {{K}ag{N}et: Knowledge-aware graph networks for commonsense reasoning}.
\newblock In \emph{EMNLP-IJCNLP}.

\bibitem[{Liu et~al.(2019{\natexlab{a}})Liu, Jiang, He, Chen, Liu, Gao, and Han}]{liu2019variance}
Liyuan Liu, Haoming Jiang, Pengcheng He, Weizhu Chen, Xiaodong Liu, Jianfeng Gao, and Jiawei Han. 2019{\natexlab{a}}.
\newblock \href {https://arxiv.org/abs/1908.03265} {On the variance of the adaptive learning rate and beyond}.
\newblock In \emph{ICLR}.

\bibitem[{Liu et~al.(2019{\natexlab{b}})Liu, Ott, Goyal, Du, Joshi, Chen, Levy, Lewis, Zettlemoyer, and Stoyanov}]{liu2019roberta}
Yinhan Liu, Myle Ott, Naman Goyal, Jingfei Du, Mandar Joshi, Danqi Chen, Omer Levy, Mike Lewis, Luke Zettlemoyer, and Veselin Stoyanov. 2019{\natexlab{b}}.
\newblock \href {https://arxiv.org/abs/1907.11692} {Roberta: A robustly optimized bert pretraining approach}.
\newblock \emph{arXiv preprint}.

\bibitem[{Mihaylov et~al.(2018)Mihaylov, Clark, Khot, and Sabharwal}]{mihaylov-etal-2018-suit}
Todor Mihaylov, Peter Clark, Tushar Khot, and Ashish Sabharwal. 2018.
\newblock \href {https://doi.org/10.18653/v1/D18-1260} {Can a suit of armor conduct electricity? a new dataset for open book question answering}.
\newblock In \emph{EMNLP}.

\bibitem[{Reimers and Gurevych(2019)}]{reimers-gurevych-2019-sentence}
Nils Reimers and Iryna Gurevych. 2019.
\newblock \href {https://doi.org/10.18653/v1/D19-1410} {Sentence-{BERT}: Sentence embeddings using {S}iamese {BERT}-networks}.
\newblock In \emph{EMNLP-IJCNLP}.

\bibitem[{Robinson and Wingate(2023)}]{robinson2022leveraging}
Joshua Robinson and David Wingate. 2023.
\newblock \href {https://openreview.net/forum?id=yKbprarjc5B} {Leveraging large language models for multiple choice question answering}.
\newblock In \emph{ICLR}.

\bibitem[{Sun et~al.(2022)Sun, Shi, Qi, and Zhang}]{sun-etal-2022-jointlk}
Yueqing Sun, Qi~Shi, Le~Qi, and Yu~Zhang. 2022.
\newblock \href {https://doi.org/10.18653/v1/2022.naacl-main.372} {{J}oint{LK}: Joint reasoning with language models and knowledge graphs for commonsense question answering}.
\newblock In \emph{NAACL}.

\bibitem[{Talmor et~al.(2019)Talmor, Herzig, Lourie, and Berant}]{talmor-etal-2019-commonsenseqa}
Alon Talmor, Jonathan Herzig, Nicholas Lourie, and Jonathan Berant. 2019.
\newblock \href {https://doi.org/10.18653/v1/N19-1421} {{C}ommonsense{QA}: A question answering challenge targeting commonsense knowledge}.
\newblock In \emph{NAACL}.

\bibitem[{Tenney et~al.(2019)Tenney, Das, and Pavlick}]{tenney-etal-2019-bert}
Ian Tenney, Dipanjan Das, and Ellie Pavlick. 2019.
\newblock \href {https://doi.org/10.18653/v1/P19-1452} {{BERT} rediscovers the classical {NLP} pipeline}.
\newblock In \emph{ACL}.

\bibitem[{Wang et~al.(2022)Wang, Zhang, Yang, Song, and Qin}]{wang2022gnn}
Kuan Wang, Yuyu Zhang, Diyi Yang, Le~Song, and Tao Qin. 2022.
\newblock \href {https://openreview.net/forum?id=hzmQ4wOnSb} {{GNN} is a counter? revisiting {GNN} for question answering}.
\newblock In \emph{ICLR}.

\bibitem[{Xu et~al.(2021)Xu, Zhu, Xu, Liu, Zeng, and Huang}]{xu-etal-2021-fusing}
Yichong Xu, Chenguang Zhu, Ruochen Xu, Yang Liu, Michael Zeng, and Xuedong Huang. 2021.
\newblock \href {https://doi.org/10.18653/v1/2021.findings-acl.102} {Fusing context into knowledge graph for commonsense question answering}.
\newblock In \emph{ACL-IJCNLP Findings}.

\bibitem[{Yasunaga et~al.(2022)Yasunaga, Bosselut, Ren, Zhang, Manning, Liang, and Leskovec}]{yasunaga2022deep}
Michihiro Yasunaga, Antoine Bosselut, Hongyu Ren, Xikun Zhang, Christopher~D Manning, Percy~S Liang, and Jure Leskovec. 2022.
\newblock \href {https://openreview.net/forum?id=4NpoSrT8uU-} {Deep bidirectional language-knowledge graph pretraining}.
\newblock In \emph{NeurIPS}.

\bibitem[{Yasunaga et~al.(2021)Yasunaga, Ren, Bosselut, Liang, and Leskovec}]{yasunaga-etal-2021-qa}
Michihiro Yasunaga, Hongyu Ren, Antoine Bosselut, Percy Liang, and Jure Leskovec. 2021.
\newblock \href {https://doi.org/10.18653/v1/2021.naacl-main.45} {{QA}-{GNN}: Reasoning with language models and knowledge graphs for question answering}.
\newblock In \emph{NAACL}.

\bibitem[{Ye et~al.(2023)Ye, Cao, Chen, Xu, and Zou}]{Ye_Cao_Chen_Xu_Zou_2023}
Qichen Ye, Bowen Cao, Nuo Chen, Weiyuan Xu, and Yuexian Zou. 2023.
\newblock \href {https://doi.org/10.1609/aaai.v37i11.26629} {{FiTs}: fine-grained two-stage training for knowledge-aware question answering}.
\newblock In \emph{AAAI}.

\bibitem[{Zhang et~al.(2020)Zhang, Zhao, Wu, Zhang, Zhou, and Zhou}]{Zhang_Zhao_Wu_Zhang_Zhou_Zhou_2020}
Shuailiang Zhang, Hai Zhao, Yuwei Wu, Zhuosheng Zhang, Xi~Zhou, and Xiang Zhou. 2020.
\newblock \href {https://doi.org/10.1609/aaai.v34i05.6502} {{DCMN+}: Dual co-matching network for multi-choice reading comprehension}.
\newblock In \emph{AAAI}.

\bibitem[{Zhang et~al.(2022)Zhang, Bosselut, Yasunaga, Ren, Liang, Manning, and Leskovec}]{DBLP:conf/iclr/0001BYRLML22}
Xikun Zhang, Antoine Bosselut, Michihiro Yasunaga, Hongyu Ren, Percy Liang, Christopher~D Manning, and Jure Leskovec. 2022.
\newblock \href {https://openreview.net/forum?id=41e9o6cQPj} {Grease{LM}: Graph {REAS}oning enhanced language models}.
\newblock In \emph{ICLR}.

\end{thebibliography}

\end{document}